\documentclass[conference]{IEEEtran}
\IEEEoverridecommandlockouts

\usepackage{cite}
\usepackage{amsmath, amssymb, amsfonts, amsthm}
\usepackage{graphicx}
\usepackage{textcomp}
\usepackage{xcolor}
\usepackage{booktabs}
\usepackage{url}
\usepackage[utf8]{inputenc}
\usepackage{mathtools}
\usepackage{bm}
\usepackage{multirow}
\usepackage[ruled,vlined,linesnumbered]{algorithm2e} 
\usepackage{hyperref}

\usepackage{amsmath,amsfonts,bm}

\def\eqref#1{equation~\ref{#1}}

\def\1{\bm{1}}

\def\vv{{\bm{v}}}

\DeclareMathAlphabet{\mathsfit}{\encodingdefault}{\sfdefault}{m}{sl}
\SetMathAlphabet{\mathsfit}{bold}{\encodingdefault}{\sfdefault}{bx}{n}

\newtheorem{theorem}{Theorem}[section]

\newtheorem{proposition}{Proposition}[section]
\newtheorem{property}{Property}[section]
\newtheorem{definition}{Definition}[section]

\begin{document}
\title{MASE: Interpretable NLP Models via Model-Agnostic Saliency Estimation}
\author{
\IEEEauthorblockN{Zhou Yang, Shunyan Luo, Jiazhen Zhu, and Fang Jin}
\IEEEauthorblockA{\textit{Department of Statistics} \\
\textit{The George Washington University}\\
Washington, DC, USA \\
zhou\_yang@gwu.edu, shine\_lsy@gwu.edu, jason.jz.zhu@gmail.com, fangjin@gwu.edu}
}

\maketitle




\vspace{-4em}
\begin{abstract}

Deep neural networks (DNNs) have made significant strides in Natural Language Processing (NLP), yet their interpretability remains elusive, particularly when evaluating their intricate decision-making processes. Traditional methods often rely on post-hoc interpretations, such as saliency maps or feature visualization, which might not be directly applicable to the discrete nature of word data in NLP. Addressing this, we introduce the Model-agnostic Saliency Estimation (MASE) framework. MASE offers local explanations for text-based predictive models without necessitating in-depth knowledge of a model's internal architecture. By leveraging Normalized Linear Gaussian Perturbations (NLGP) on the embedding layer instead of raw word inputs, MASE efficiently estimates input saliency. Our results indicate MASE's superiority over other model-agnostic interpretation methods, especially in terms of Delta Accuracy, positioning it as a promising tool for elucidating the operations of text-based models in NLP.
\end{abstract}



\maketitle

\section{Introduction}
\label{intro}
Deep learning models are becoming increasingly prevalent in Natural Language Processing (NLP) systems. Various efficient deep neural networks (DNNs) have been proposed, ranging from classic models such as the Recurrent Neural Network (RNN)~\cite{yin2017comparative} and Long-Short Term Memory (LSTM)~\cite{hochreiter1997long} to more recent Bidirectional Encoder Representations from Transformers (BERT)~\cite{devlin2018bert}. Although these DNNs have shown great success in natural language processing tasks, their lack of explainability and/or interpretability remains a major weakness when comparing and deploying these black-box models for NLP tasks such as text classification and resume recommendation. Moreover, as NLP models become increasingly complex with millions or even billions of parameters, explaining the precise decision-making process of a DNN is becoming increasingly challenging.
There is thus an urgent need to establish reliable interpretations for text-based predictive models in order to build trust and transparency between users and the models we rely on in our everyday lives~\cite{lipton2018mythos, guo2024bias, guo2024transparent, wu2023driver}. 

There have been a number of efforts to explain black-box models in NLP tasks and reveal how they work. One popular approach is to create a complex but transparent model to obtain a promising interpretation~\cite{ustun2016supersparse, yang2022tutorial}, 
while the other well-established one is to adopt post-hoc interpretation methods to assess the deep learning models. 
These post-hoc interpretation methods often rely on the concept of a saliency map to represent the importance of the input variables for the final decision, approximating the importance in terms of the gradient of output score with respect to the inputs~\cite{simonyan2014deep, su2024explanation}. 
An alternative post-hoc approach is to compute either the amount that inputs contribute to the output score~\cite{montavon2017explaining} or their Shapley value~\cite{lundberg2017unified}. These methods were initially proposed to interpret convolutional neural networks but are now being applied to NLP models to study word contribution~\cite{li2015visualizing}. Rather than using word-level explanations, several approaches focus on feature visualization by investigating the pattern that the hidden neurons of a model are trying to detect~\cite{olah2018building}. 
The key idea here is to iteratively update a randomly initialized input to explore a specific behavior in the hidden layers of the model, for example, by maximizing either the activation values of neurons or the score of a class. The resulting input can then be visualized as a local interpretation of what the deep learning model expects to `see'.
However, such techniques cannot be directly applied to word-level data in NLP models since word representations are discrete and the meaning cannot be abstracted efficiently, making the resulting input, which is a sequence of abstract vector representations, difficult to interpret with respect to the original text. 
Moreover, it is not clear whether individual words should be the primary unit of attribution. The meaning of each word is heavily entangled with other words and is thus far removed from high-level concepts like the output class. 
Therefore, the utility of individual word/feature-based visualization techniques to explain deep learning models is very limited.

\begin{figure}
    \centering
\includegraphics[width=0.85\linewidth]{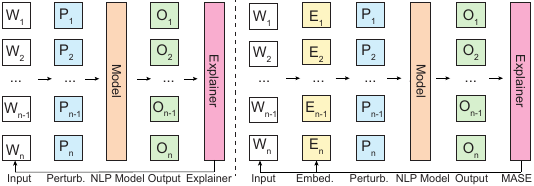}
    \caption{Pipelines of MASE vs. other perturbation-based methods. Left: Pipeline of other perturbation-based methods. Perturbations are often made at the word level and use a special token to represent the absence of words. Right: Pipeline of MASE. We perturb the embedding of words instead of the raw words so that the perturbation space is largely expanded from binary to Euclidean, which is more precise to describe the local behavior of the target model. }
    \label{fig:FrameMASE}
\end{figure}

The fact that explaining deep neural networks in an NLP setting remains challenging is primarily due to the unique data structure of the inputs involved and the corresponding subjective evaluations. Pixel intensities and audio frequency are often used as raw input on an image or audio classification, while the text is either tokenized or embedded before being fed into these deep models. Traditional local interpretation methods are largely based on iteratively masking each word in turn in a series of words to study the contribution of the words that have been masked~\cite{ribeiro2016should}. Given an input of $N$ words, the corresponding perturbation lies in binary space, which contains a total of $2^N$ possibilities. This exponential increase in sample space of masking is computationally expensive, especially considering the high-dimensional input in a typical NLP scenario, and there is no guarantee that the masked text and the original text will even stay in the same neighborhood after embedding layers. Moreover, the choice of alternatives for the masking word becomes a hyper-parameter for interpretation methods based on word-level analyses, which brings uncertainty and inconsistency in the explanatory results. A similar uncertainty, known as Homoscedastic uncertainty~\cite{zheng2022multi,xu2022inferring}, can be seen during the training stage. Homoscedastic uncertainty is the uncertainty that remains the same right across the input space when a special token, such as `[CLS]' or `[SEP]', is used to replace a targeted word. In contrast, Heteroscedastic uncertainty~\cite{zheng2022multi,xu2022inferring} is an input-dependent uncertainty that is incurred when a short text is padded into a fixed length with special tokens, but the effects of these special tokens are unclear. Discrete data structures and uncertainties in the training stage increase the challenges for efforts to explain NLP deep learning models.

To overcome the aforementioned challenges, in this project, we introduce a new perturbation-based interpretation technique, model-agnostic saliency estimation (MASE), to create local interpretations of arbitrary deep models using the embedding layer in NLP tasks. The framework of our proposed MASE method is shown in Figure~\ref{fig:FrameMASE}. Unlike existing model-agnostic methods, which perform local interpretation utilizing word-level perturbations, here we opted to apply the perturbation to the embedding representation and then map the saliency back onto the original text. The perturbation is implemented with Normalized Linear Gaussian Perturbation (NLGP), as this enables us to compute local gradients in any direction and is not limited to the 0-1 direction that restricts word-level perturbation. As a consequence, this novel perturbation scheme is able to generate meaningful perturbations while at the same time retaining the word embedding. 
Our proposed MASE method thus approximates the saliency based on the embedding-level perturbation more accurately while eliminating the uncertainties that arise as a result of the choice of replacement words. Furthermore, a Lasso-penalized version of MASE has been designed to achieve a faster convergence rate since the feasible region is reduced considerably by the penalizing constraint. Last but not least, we present a quantitative and general evaluation metric, ``Delta Accuracy'', that works in a real-world setting to measure the performance of different explanatory methods. The proposed metric excels due to the fact that it works without the need to use a ``human-in-the-loop'' evaluation to manually check the validity of explanatory algorithms, thus filling the major gap that arises because the evaluation techniques currently used in explainable learning are still immature, with a primary focus on human-in-the-loop evaluations, and quantitative general evaluation schemes are yet to be explored~\cite{das2020opportunities, nguyen2019spatial, du2019twitter}. 

The contributions of this paper are summarized as follows.
\begin{itemize}
    \item We propose a constrained perturbation-based technique to explain how a text-based predictive model works in black-box settings. The proposed method achieves excellent performance, eliminating the uncertainty caused by masking alternatives and producing consistent explanations for the results. 
    \item We present a new perturbation scheme for saliency estimation in embedding space. In contrast with the standard Gaussian perturbation, our proposed new perturbation scheme preserves the embedding direction for each word and balances the correlations between words. 
    \item We theoretically prove that MASE achieves the best performance measure in terms of infidelity measure and demonstrate that the MASE is a unified framework that connects LIME~\cite{ribeiro2016should}, SHAP~\cite{lundberg2017unified}, Occlusion~\cite{zeiler2014visualizing}, and IG~\cite{sundararajan2017axiomatic}.
\end{itemize}

\section{Background and Related Work}
Interpretation is essential for building trust in modern deep models, yet their complexity demands clarification of key concepts to establish a solid foundation for interpretable learning.

An interpretation is the output of a method that reveals how a model makes decisions. While linear regression weights directly reflect feature importance, understanding deep models such as CNNs is far more challenging due to their numerous parameters and nonlinear activations. This motivates methods that generate interpretable explanations—by identifying key feature combinations~\cite{ribeiro2016should} or tracing influential training samples~\cite{koh2017understanding}.
In contrast, model interpretability measures how understandable a model’s reasoning is to humans~\cite{doshi2017towards,lage2019evaluation}, a subjective but crucial notion. Most interpretation studies focus on computer vision~\cite{yuan2019interpreting}; the most relevant prior work, Linearly Estimated Gradient (LEG)~\cite{luo2021statistically}, provides a statistically consistent saliency estimator for CNNs.

Existing interpretation methods fall into two categories: model-specific and model-agnostic.
Model-specific methods rely on knowledge of the model architecture, such as neuron-level visualizations that quantify feature contributions or maximize activations~\cite{erhan2009visualizing}.
Model-agnostic methods treat the model as a black box. Examples include LIME~\cite{ribeiro2016should}, which fits a sparse linear surrogate on perturbed samples, and SHAP~\cite{lundberg2017unified,chen2018lshapley}, which combines Shapley values with additive feature attribution for consistency with methods like LRP, LIME, and DeepLIFT~\cite{shrikumar2017learning}.
Building on these ideas, we estimate partial derivatives of input features over a continuous probability distribution rather than fixed points, improving fidelity and sensitivity~\cite{yeh2019fidelity}.
Although model-agnostic methods are generally more consistent and generalizable, they are computationally heavier due to sampling.
This study focuses on model-agnostic interpretation and provides theoretical analysis of sample size to balance accuracy and efficiency.

Across both paradigms, saliency maps are the most common explanation form~\cite{du2018towards}. They assign importance scores to input elements, often visualized as heatmaps. Gradient-based methods compute these scores via derivatives of predictions with respect to inputs, while perturbation-based methods assess output changes from modified inputs~\cite{ribeiro2016should}.
Representative gradient-based methods include Vanilla Gradient~\cite{simonyan2013deep}, Gradients × Input~\cite{shrikumar2017learning}, and Grad-CAM~\cite{selvaraju2017grad}.
In this work, we use calibrated perturbations to estimate local gradients in deep NLP models. By combining strengths of perturbation- and gradient-based strategies, our proposed MASE framework offers faithful, model-agnostic interpretations of local behaviors in neural NLP predictions.

\section{Method}
\label{method}
In this section, we first introduce the definition of saliency and then discuss our proposed estimand of approximating saliency in a text setting. Furthermore, we present the proposed MASE framework, provide theoretical proof of its properties, and show the efficient computation algorithms for our estimand. Lastly, we discuss the connection between the proposed model and existing methods. 

The explanation in this work is defined as a meta learner that is a function of a function. Consider the following general predictive learning setting: given the input space $\mathcal{X} \subseteq \mathbb{R}^d$, an output space $\mathcal{Y} \subseteq \mathbb{R}$, and a black-box function $\mathbf{f}: \mathbb{R}^d \mapsto \mathbb{R}$, which predicts the output $\mathbf{f}(\mathbf{x_0})$ at an arbitrary test input $\mathbf{x_0} \in \mathbb{R}^d$. Then a feature attribution explanation is a function $\gamma: \mathcal{F} \times \mathbb{R}^d \mapsto \mathbb{R}^d$, that given a black-box predictor $\mathbf{f}$, and a test point $\mathbf{x}$, provides importance scores $\gamma(\mathbf{f}, \mathbf{x})$ for the set of input features.

\subsection{Saliency Estimation}
Saliency estimation is achieved by recovering the gradient of the deep learning model with regard to the input. For instance, given an image $x \in \mathcal{X}$ and a model $f:\mathcal{X} \rightarrow \mathcal{R}$, the model output $f(x) = [f_1(x), \dots, f_C(x)]$ for $C$ classes, the ``Image-Specific Class Saliency''~\cite{simonyan2013deep} of single pixel $x_{i,j}$ can be written as:
\begin{equation}
    \textit{Saliency}({x_{i,j}}) = \frac{\textbf{d}f_c(x)}{\textbf{d}x_{i,j}}.
\end{equation}

Similarly, we can define saliency in NLP settings with respect to the embeddings. Although saliency is a good interpretation approach to reveal the influential parts of input for the deep model prediction, the saliency is noisy with the high dimensionality of the image data. Thus an average gradient around $x$ is preferred to obtain smoother results.

\subsection{ Model-Agnostic Saliency Estimation}
\label{MASE}
Saliency is well-defined in computer vision since the pixel intensity can be treated as a continuous variable ranging from 0 to 1. However, the continuity assumption defined on the raw input is no longer satisfied in natural language processing. Existing methods claim that each word in the input has two states (``exist'' or ``absent'') and compute the word importance by making perturbations on the state space (i.e. randomly selecting some words and replacing them with an alternative token to represent their absence). Such a scheme can not be applied to the approximation of saliency because 1) there is so far no study verifying the consistency of these interpretations due to the different choice of alternatives; 2) such perturbations are only based on a human heuristic, which may not be treated as the neighbor of the original text by the target model. 

Motivated by the embedding representations built in most deep NLP models, we propose a new saliency estimation framework that is able to address the aforementioned two concerns using perturbation on the raw words. Specifically, let $f$ denote the deep learning model to be explained, and $f(x)$ denote the output given input text $x=(w_1,w_2,\dots, w_n)$ that consists of $n$ words ($w_1,w_2,\dots, w_n$). The embedding representation of $x=(w_1,w_2,\dots, w_n)$ can be denoted as $e(x)$. 
Suppose an m-dimensional representation is generated for each word $w_{i}$ after the embedding layer (i.e. $\forall w_{i}, e(w_{i}) = (a_{i1},a_{i2},...,a_{im}) \in \mathbb{R}^m)$, it is natural to define saliency within the Euclidean space spanned by the embedding vector. Furthermore, meaningful perturbations can be applied through the embedding space to produce a reliable saliency estimate as local interpretation to explain the importance of each word in terms of the target deep learner. As a result, we define a novel estimate of saliency in Definition \ref{def:MASE}.


\begin{definition}[Model-Agnostic Saliency Estimation (MASE)]
\label{def:MASE}
Let $f: \mathcal{X} \rightarrow [0,1]$ denote a deep learning model with an embedding layer, and let the input sequence be $x_0 = (w_1, w_2, \dots, w_n)$ consisting of $n$ tokens. The model output $f(e(x))$ represents the predicted probability of a target class, where each token embedding is $e(w_i) = (a_{i1}, a_{i2}, \dots, a_{im}) \in \mathbb{R}^m$.

Let $F_{\mathcal{M}}$ be a probability distribution defined over the embedding space $\mathcal{M} \subset \mathbb{R}^m$, and let $\mathcal{L}$ be a loss function measuring the discrepancy between model responses and their local linear approximations. Then, the \textit{Model-Agnostic Saliency Estimate} (MASE) is defined as
\begin{equation}
\label{eq:definition_1}
\begin{aligned}
\gamma(f, x_0, F_{\mathcal{M}}) 
&= \arg\min_{g} 
\mathbb{E}_{e(x) \sim F_{\mathcal{M}}}
\Big[
\mathcal{L}\big(
f(x) - f(x_0), \\
&\hspace{5.5em}
g^{\top}\!\big(e(x_0) - e(x)\big)
\big)
\Big],
\end{aligned}
\end{equation}
where ${g} \in \mathbb{R}^n$ is the decision variable of $\gamma\left(f, {x_{0}}, F\right)$.
\end{definition}

MASE is based on the smoothing gradient on the embedding vectors created by the target model around the point of interest $x_0$. It gives the best linear approximation of the average local gradient in terms of specified loss function over the embedding space $\mathcal{M}$. 
As a perturbation-based interpretation method, the change of perturbation space from raw text to embedding space allows us to define saliency successfully over a compact space and expand the flexibility to choose meaningful perturbations.
Furthermore, MASE takes the expectation over the space $\mathcal{M}$, which belongs to the class of model interpretation techniques using local smoothing. Such a technique is known to defend adversarial manipulations reliably~\cite{dombrowski2019explanations}, be more faithful to the model~\cite{yeh2019fidelity}, provide visualization more consistent to human knowledge, and perform better on the sensitivity analysis. 

We note that the choice of space $\mathcal{M}$ and the variance of $F$ are critical for MASE. As 
if $F$ has a high variance, then samples from $e(x_0) + F$ are substantially different from
$e(x_0)$, and MASE might no longer be accurate in explaining the model at point $x_0$. At this time, the restricted subspace $\mathcal{M}$
can prevent such failure with dedicated design. To be specific, this constraint benefits the MASE framework in two ways. First, the space restriction on the perturbation distribution $F$ makes the estimate more accurate and reliable since the training data are texts, which are highly structured based on unique concepts like grammar and semantics, compared to other types of data. Second, $\gamma(f, {x_{0}}, F_{\mathcal{M}} )$ can take values in any direction of the neighborhood of ${x_0}$ in the embedding space and is not limited to the binary cases in the masking scheme such that ${x} \in \{0,1\}^n$. On the contrary, samples generated without space restriction by completely random perturbations lack semantic meaning and are challenging to classify even by humans. Thus, it may perform poorly and cause a large bias in the approximated saliency. Therefore, it is more plausible to make perturbations on a reasonable subspace over the whole space. 

\begin{figure}
    \centering
\includegraphics[width=\linewidth]{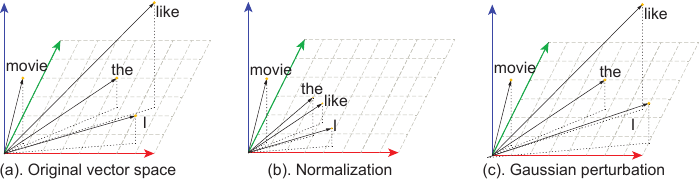}
    \caption{Illustration of Normalized Linear Gaussian Perturbation Procedure in 3-dimensional Cartesian Space. (a) shows the original words of the toy example and their corresponding word embedding vectors in 3-D. (b) illustrates the normalized embedding vectors. (c) displays the embedding vectors after applying the Gaussian perturbation.}
    \label{fig:NLGP}

\end{figure}

\subsection{Normalized Linear Gaussian Perturbation}
\label{GIP}
As discussed above, how to define the subspace $\mathcal{M}$ and corresponding distribution $F$ is critically important. Ideally, the perfect subspace should include all possible sentences with similar semantics and proper grammar and should be close to $x_0$ based on certain pre-defined distance matrices. In addition, it cannot be too small to miss any significant directions. Considering the complexity and flexibility of the semantic explanation, it is challenging to describe the perfect subspace. Existing perturbation-based explanations make perturbation at the word level~\cite{ribeiro2016should}, e.g., replacing $w_{i}$ by a fixed token. However, it is not guaranteed that the perturbed text will be in the neighborhood of the original text. We notice that the weights of the embedding layers are learnable during the training stage and hence the embedded word vector shall contain hidden information to some degree. Therefore, we define $\mathcal{M} = \{(b_1,b_2,\dots,b_n): \arccos \frac{{a_i}\cdot {b_i}}{|{a_i}||{b_i}|}\leq \theta,~ \forall i \}$. Geometrically speaking, we restrict the perturbation on the cone centered with each individual embedded word vector and the angle between the perturbed vector and the original one is less than $\theta$. To maximize the restriction on $\mathcal{M}$, 
we design a new perturbation scheme called Normalized Linear Gaussian Perturbation (NLGP) in Definition~\ref{def:GIP} by setting $\theta \rightarrow 0$, which forces the perturbed vector to share exactly the same direction with the original word vector. In this way, the perturbation implementation can be simplified substantially. We also apply normalization on word vectors before adding Gaussian noise to balance the different effects among all word vectors. 
\begin{definition}[Normalized Linear Gaussian Perturbation]
\label{def:GIP}
  Let instance $x = ({w}_1, {w}_{2}, \cdots, {w}_{n})$ be a text sequence, where $w_i$ indicates single word or symbol. In the embedding space, $w_i$ can be represented by a vector with an equal length of $m$, where $m$ is determined by the model. The Normalized Linear Gaussian Perturbations (NLGP) of $x$ can be written as: $X' = \{x_{c}: x_c ={c}^{T}e({x}), {c} = (c_1,c_2,\dots,c_n) \in \mathbb{R}^n, {c} \sim \text{Gaussian}(\textbf{1}, \Sigma) \}$
\end{definition}
Definition~\ref{def:GIP} shows how we perturb the embedding representations. That is, ${c}$ is sampled from a Gaussian distribution $Gaussian(\textbf{1}, \Sigma)$, and the perturbation results of $e(x)$ is $x_c = {c}^{T}e({x})$. Figure~\ref{fig:NLGP} shows an illustration of the NLGP scheme in a 3-dimensional setting.
We treat each embedding vector as one element and make Gaussian perturbation on the weight of the embedding vector, which preserves the row space spanned by the input $x_0$ under such a perturbation.
This space invariant property enables MASE to better leverage the local saliency of each word while keeping the semantic impact. To be more concrete, MASE perturbation can be seen as dilation and contraction on the polyhedron spanned by the embedding vectors. In addition, MASE takes each embedding vector as one element, which is equivalent to applying normalization on each embedding vector. In this way, MASE generates fair comparisons among words and decreases the impact of the length of different vectors. Moreover, Theorem~\ref{theorem:1} shows that the MASE estimate can be computed empirically if the covariance matrix of the perturbation is positive definite. Such a characteristic provides a convenient way to calculate the MASE estimator since we can choose $\Sigma$ to make sure it is positive definite.

\begin{theorem}
\label{theorem:1}
Let $x_0$ be an input sequence of $n$ tokens to a deep learning model $f$, and let $Z$ be a random variable in the embedding space with a centered distribution $F$, i.e., $Z \sim F$ and $\mathbb{E}[Z] = \mathbf{0}_n$. 
Assume that the covariance of $\operatorname{vec}(Z)$ exists and is positive definite. 
Let $\Sigma = \operatorname{Cov}(\operatorname{vec}(Z))$. 
Then, the model-agnostic saliency estimator $\gamma(f, x_0, F)$ has the following closed-form expression:
\begin{equation}
\label{eq:lemma1}
\begin{aligned}
\gamma(f, x_0, F)
&= \operatorname{vec}^{-1}\!\Big(
\Sigma^{-1} \,
\mathbb{E}_{Z \sim F}\Big[
\big(f(e(x_0 + Z)) - f(e(x_0))\big)  \\
&\hspace{8em}\times
\operatorname{vec}(Z)
\Big]
\Big).
\end{aligned}
\end{equation}

\end{theorem}
Equation~\ref{eq:lemma1} can be derived by taking the derivative of Equation~\ref{eq:definition_1} with respect to $g$ and setting it to zero. Note that square error is specified for the loss function. 
Next, we will introduce Infidelity score~\cite{yeh2019fidelity}, a general evaluation of the explanation method, and show the proposed MASE framework achieves the optimal explanation in terms of the infidelity measure.
The infidelity measure is defined as the average difference between the dot product of the output perturbation ($f(x)-f(x_0)$) and the input perturbation ($x-x_0$) to the explanation. 
 
 \begin{definition}[Infidelity]
 \label{def:infidelity}
  Given a black-box function ${f}$, explanation functional $\gamma$, a random variable ${Z} \in \mathbb{R}^d$ is sampled from the distribution $F$, which represents meaningful distributions of interest, the explanation infidelity of $\gamma$ is defined as~\cite{yeh2019fidelity}:
$$
\operatorname{IF}(\gamma, {f}, {x})=\mathbb{E}_{{Z} \sim {{F}}}\left[\left({Z}^T \gamma({f,x})-({f}({x})-{f}({x}-{Z}))\right)^2\right],
$$
where $z$ represents significant perturbations around ${x_0}$, and can be specified in various ways. For instance, it can be built based on the difference to noisy baseline: ${z}={x}-{x}_0$. The perturbed input ${x}={x}_0+z$, for some zero mean random vector $z$, for instance, $z \sim \mathcal{N}\left(0, \sigma^2\right)$
 \end{definition}

Given the measure of infidelity, a natural intuition is to find the optimal explanation that minimizes infidelity. This optimal explanation has a closed form given some assumptions and is illustrated in the following theorem.

\begin{theorem}
\label{th:connectIG}
Let the perturbation $Z$ be a random variable defined on the probability space $(\Omega, \mathcal{F}, \mu)$ such that 
$\int ZZ^{\top} \, d\mu(z)$ is invertible, i.e., its inverse 
$\big(\int ZZ^{\top} \, d\mu(z)\big)^{-1}$ exists. 
Define the integrated gradient of $f(\cdot)$ between $(x - Z)$ and $x$ as
\[
\operatorname{IG}(f, x, Z)
= \int_{0}^{1} \nabla f(x + (t - 1)Z)\, dt,
\]
which may be replaced by any functional satisfying 
$Z^{\top}\!\operatorname{IG}(f, x, Z)
= f(x) - f(x - Z)$~\cite{sundararajan2017axiomatic}.
Then the following statements hold:
\begin{enumerate}
    \item The optimal explanation $\gamma^{*}(f, x)$ that minimizes the infidelity for perturbations $Z$ is
    \begin{equation}
    \label{eq:optimal_explanation}
    \gamma^{*}(f, x)
    = \!\left(\!\int ZZ^{\top} d\mu(z)\!\right)^{\!-1}
    \!\left(\!\int ZZ^{\top} \operatorname{IG}(f, x, Z) \, d\mu(z)\!\right).
    \end{equation}
    \item The MASE estimator derived in Theorem~\ref{theorem:1} attains this optimal solution and thus minimizes the infidelity measure.
\end{enumerate}
\end{theorem}

The SmoothGrad in Thoerem~\ref{th:connectIG} can be written as $$\gamma_k({f}, {x}):=\left[\int_{{z}} k({x}, {z})\right]^{-1} \int_{{z}} \gamma({f}, {z}) k({x}, {z}) d \mu_{(z)}$$ where the Gaussian kernel $k(x,z)$ can be replaced by arbitrary kernel function. Therefore, the optimal solution of Theorem~\ref{th:connectIG} can be viewed as applying a smoothing operation that is similar to SmoothGrad on Integrated Gradients, where a special kernel ${ZZ}^T$ is used instead of the original kernel $k({x}, {z})$.

\subsection{Property of the MASE explanation}
Since MASE in essence uses first-order Tylor approximation to estimate the unknown black-box model, it belongs to the additive model family~\cite{hastie2017generalized}. A preferable attribute of the class of additive feature attribution methods is the presence of several desirable properties. 

The first desirable property is local accuracy. This property states that a black-box model $f(x)$ can be well-approximated at the neighborhood of $x$ by the explanation method $\gamma(x\prime, f, F)$, where $x\prime$ corresponds to the original input $x$. 

\begin{property}[Local Accuracy]
$f(x_0)=\gamma\left(x^{\prime}\right)=\gamma_0+\sum_{i=1}^M \gamma_i x_i^{\prime}$
The explanation model $\gamma\left(x^{\prime}\right)$ matches the original model $f(x)$ when $x\prime$ is sampled from the near-by of the original $x$.
\end{property}

In certain instances, some features might be absent or represented as zeros in the input data. If the simplified inputs are indicative of the presence of features, then it's essential that features absent in the original input do not influence the explanation. The second property of LEMA framework is missingness.

\begin{property}[Missingness]
If a feature $x_i^{\prime}=0$, indicating the absence or no observation, the explanation $\gamma_i=0$, which enforces that features for which $x_i^{\prime}=0$ should be attributed with no impact.
\end{property}

The third attribute is named as consistency. Consistency states that if perturbations are sufficiently minuscule, the approximated saliency estimator $\gamma_\epsilon^*(f,x)$ will approximate the gradient of the black box that needs explanation.

\begin{property}
Assume that the perturbation is denoted by ${Z}_\epsilon=\epsilon \cdot {e}_i$ such that $x\prime -x = Z_{\epsilon}$, where ${e}_i$ represents a coordinate basis vector. In this case, the optimal explanation $\gamma\epsilon^*({f}, {x}, Z_{\epsilon})$, as it pertains to infidelity for perturbations ${Z}_{\epsilon}$, adheres to the condition $\lim_{\epsilon \rightarrow 0} \gamma_{Z^*}({f}, {x})=\nabla {f}({x})$.
\end{property}

\subsection{Efficient Estimate of MASE and its Consistency}
\label{MASE_Sparse}
In NLP classification, given a whole paragraph, people often make their decisions based on a few critical sentences or words. Such an idea is often implemented through an attention mechanism or other similar techniques. Hence we propose a variant of MASE, called MASE\_Sparse, to achieve a higher convergence rate and sparse interpretation by introducing $L_1$ regularization in Definition~\ref{def:MASE_Sparse}.

\begin{definition}[\textbf{Sparse Model-Agnostic Saliency Estimation (MASE\_Sparse)}]
\label{def:MASE_Sparse}
Let $L \ge 0$ be a sparsity-controlling hyperparameter.  
The $L_1$-constrained MASE estimator is defined as
\begin{equation}
\label{eq:definition_2}
\begin{aligned}
\gamma(f, x_0, F_{\mathcal M})
&= \arg\min_{g \in \mathbb R^n}
\mathbb E_{x \sim F_{\mathcal M}}
\Big[
f(x)-f(x_0) \\
&\hspace{6em}
-\,\langle g,\,e(x_0)-e(x)\rangle
\Big]^2.
\end{aligned}
\end{equation}

subject to the sparsity constraint
\begin{equation*}
\|g\|_1 \le L,
\end{equation*}
where $g \in \mathbb R^n$ denotes the local saliency vector.
\end{definition}

The level of $L$ controls the sparsity of the corresponding solution. We also notice that the Equation~\ref{eq:definition_2} can be re-written as 
an alternative formula using a high probability approach~\cite{fan2013features, luo2021statistically}.

\begin{equation}
\label{eqn:lp}
\begin{array}{l}
\gamma(f, x_{0}, F_{\mathcal{M} })=\underset{g}{\arg min}~\|g\|_{1} \\
\text { subject to }\left\|\frac{1}{N} \sum_{i=1}^{N}\left(\tilde{y}_{i} z_{i}\right)-\Sigma g\right\|_{\infty} \leq L .
\end{array}
\end{equation}
where $\tilde{y}_{i}= f(x)-f(x_0)$ and $ z_{i}=e(x)-e(x_0)$.
Equation~\ref{eqn:lp} is a linear program and can be solved much faster compared with equation~\ref{eq:definition_2} using simplex solvers.
In addition, MASE\_Sparse is statistically consistent with a convergence rate $\sqrt{\log{n}/N}$~\cite{luo2021statistically}. This implies that we can recover the true parameter as the number of model evaluations increases. Furthermore, our result proves that ignoring the log terms, one needs $ O(s \left(n\right)^{1/2})$ many model evaluations to recover the optimal $\gamma^*$.
We note that our procedure identifies the MASE coefficient up to a mean shift parameter, which is the average of the true MASE coefficient $\gamma$. However, in our numerical studies, we see that this mean shift is almost non-existent: MASE\_Sparse yields solutions that have no mean differences with the MASE coefficient, which is defined as the solution of the empirical version as $N$ is statistically large enough~\cite{altman1995statistics}. 

Algorithm~\ref{alg:MASE} illustrates the key procedures of the MASE framework. Specifically, we extract the embedding layer of the target model and determine the noise level of NLGP (we use 0.1 in the following experiment). After gathering the perturbations and corresponding prediction probabilities, we choose either Equation~\ref{eq:lemma1} or Equation~\ref{eqn:lp} to compute MASE interpretation depending on the sparsity assumption. In practice, local interpretation of a single prediction using Algorithm~\ref{alg:MASE} takes only a few seconds, which enables us to inspect every prediction in real-time.

\begin{algorithm}[t]
\caption{Model-understandable Saliency Estimation}
\label{alg:MASE}
\SetAlgoLined
\KwData{Text to be explained: $\mathcal{T}$}
\KwResult{Empirical saliency estimation $\gamma$ }
\tcp{Target model: $M$; prediction score: $P_{M,T}$; number of samples:$S$.}
Extracting embedding token \textsc{$E_{M,T}$}.\\
Specify $\Sigma$ of the Gaussian distribution in NLGP.\\
\For{$i\leftarrow 1$ \KwTo $S$}{
    compute perturbation $E^{i}_{M,T}$'s.\\
    Feed {$\{E^{i}_{M,T}\}$} to M and store outcomes {$P^i_{M,T}$}'s.\\
    With assumption on sparsity, specify the hyperparameter $L$ and compute saliency $\gamma$ by solving the linear optimization in Equation~\ref{eqn:lp}.\\
    Without assumption on sparsity, compute empirical saliency $\gamma$ by Equation~\ref{eq:lemma1}.\\
}
\end{algorithm}

Next, we demonstrate that the MASE framework can be seen as a unifying framework for many recently proposed explanations in Section~\ref{sec:connect}. We show that existing explanation methods can be viewed as optimizing the aforementioned MASE framework for varying perturbations. Moreover, MASE can also be viewed as a way to design new explanations and evaluate any existing explanations. 

\subsection{Connection to Existing Explanations}
\label{sec:connect}

Most existing techniques make different perturbations on the word-level space, while MASE is defined on a higher dimensional embedding space. However, we can still unify these existing deep learning interpretation techniques using MASE by setting $e(x) = x$ and making the perturbation distribution $F$ over the word-level space in Equation~\ref{eq:definition_3}.
\begin{equation}
\label{eq:definition_3}
    \begin{split}
        \gamma(f, {x_{0}}, F_{\mathcal{M}})=\arg \min _{g} \mathbb{E}_{{x} \sim F_{\mathcal{M}}  } \mathcal{L}(f({x})-f({x_{0}}),
        {g}^{T}({x_{0}}-{x})),
    \end{split}
\end{equation} 
where $M$ represents the binary space $\{0,1\}^n$.

\begin{proposition}
 Suppose the perturbation is $Z=U \odot x$, where $U$ is sampled from a multivariate Bernoulli distribution.
 Let the loss function $\mathcal{L}$ be a weighted square loss with kernel $\pi(x,x') = exp(-D(x,x')^2/\sigma^2)$ with $L_1$ norm penalty on the estimand, where $\sigma^2$ is a width constant and function $D$ is the cosine distance. Then the optimal explanation $\gamma(f,x)^*$ corresponding to Equation~\ref{eq:definition_3} is equivalent to the optimal explanation from LIME~\cite{ribeiro2016should}.
\end{proposition}

\begin{proposition}
 Given a customized kernel $\pi(x') = \frac{(n-1)}{{n \choose |x'|}|x'|(n-|x'|)}$ to both the loss function and sampling scheme in Equation~\ref{eq:definition_3}, where $|\cdot|$ is the number of words that are present. Moreover, if we replace $n-|x'|$ words with some random alternatives from a subset of the training dataset, then the optimal explanation is equivalent to the optimal explanation from kernel SHAP~\cite{lundberg2017unified}.
\end{proposition}

\begin{proposition}
Let ${Z}={x}-{x}_0$ be the deterministic perturbation that is the difference between ${x}$ and some baseline ${x}_0$.
Let $\gamma^*({f}, {x})$ be any explanation that is optimal with respect to infidelity for perturbations ${Z}$. Then $\gamma^*({f}, {x}) \odot {Z}$ satisfies $\sum_{j=1}^d\left[\gamma^*({f}, {x}) \odot {Z}\right]_j={f}({x})-{f}({x}-{Z})$.
\end{proposition}

\begin{proposition}
Suppose the perturbation is given by ${Z}={e}_i \odot {x}$, where ${e}_i$ is a coordinate basis vector. Let $\gamma^*({f}, {x})$ be the optimal explanation with respect to infidelity for perturbations ${Z}$. Then $\gamma^*({f}, {x}) \odot {x}$ is the occlusion explanation~\cite{zeiler2014visualizing}.    
\end{proposition}


\section{Experiment}
\label{sec:experiment}

In this section, we first present the target models, datasets, and baselines utilized for our interpretation method, and then illustrate how the proposed MASE method is evaluated using corresponding experimental settings.

\subsection{Target Models}
\label{sec:dataset}
MASE is a model-agnostic explaining method as it only requires the embedded features of the input. This model-agnostic property of the proposed method implies its powerful generalization capability for a wide range of NLP models. Two deep NLP models, LSTM and BERT, were trained respectively to exhaustively evaluate the effectiveness and performance of the MASE framework. To further investigate the performance of MASE on a specific type of network, we trained LSTM on IMDB and Reuters datasets, respectively. The specification of datasets and deep learning models are summarized in Table~\ref{tab:dataset_stats}.

{\setlength\doublerulesep{0.8pt}
\begin{table}[t]
  \centering
  \setlength{\tabcolsep}{3mm}{
  \begin{tabular}{cllrr}
    \toprule[1pt] \midrule
     \textbf{Model} & \multicolumn{1}{c}{\textbf{Dataset}} & \multicolumn{1}{c}{\textbf{Description}} & \multicolumn{1}{c}{\textbf{Datasize}} & \multicolumn{1}{c}{\textbf{\#class}} \\
    \midrule
    \multirow{2}[2]{*}{\textbf{LSTM}} &IMDB  & Movie review  & 50,000 & 2 \\
              & Reuters & newswires & 11,228 &46\\
    \midrule
    \multirow{1}[1]{*}{\textbf{BERT}} &IMDB  & Movie review  & 50,000 & 2 \\

    \bottomrule[1pt]
  \end{tabular}}
    \caption{Descriptions and statistics of IMDB and Reuters datasets used on target models.}
  \label{tab:dataset_stats}%
  \vspace{-3mm}
\end{table}%

\begin{itemize}
    \item \textbf{LSTM}, or Long short-term memory, is a classic recurrent neural network applied in NLP tasks and is capable of learning long-term dependencies, which makes it widely used for various NLP tasks. 
   In our experiments, we trained two LSTM models from scratch on IMDB and Reuters datasets separately. We fixed the model input length to 80 words and trained both models using the Adam optimizer with a learning rate of 0.001.
   All the model weights were randomly initialized during the training.
   We utilized half reviews in the IMDB dataset for training and the remaining half for testing. Our experiment showed LSTM model achieved a testing accuracy of 81.0\% on the IMDB dataset. Similar training and testing were conducted on the Reuters dataset and LSTM achieved an average testing accuracy of 84.2\%. 
    
    \item \textbf{BERT}~\cite{devlin2018bert} stands for Bidirectional Encoder Representations from Transformers. BERT randomly masks words in the sentence and then tries to predict them. Unlike the previous language models, it takes both the previous and next tokens into account at the same time. In this study, we utilized the pre-trained BERT uncased from Google~\cite{pretrainedbert2021} with 12 layers, a hidden size of 768, and 12 attention heads. 
    Built on top of the pre-trained BERT model with a fully connected layer, the model was fine-tuned on IMDB with a testing accuracy of 89.3\%. After training the BERT-based classifier, we extracted the output of the embedding layer for our perturbation. The perturbed embedding would then be fed into the BERT model.
    \item \textbf{IMDB}~\cite{maas-EtAl:2011:ACL-HLT2011} is a dataset for binary sentiment classification containing substantially more data than previous benchmark datasets. It provides a set of 25,000 highly polar movie reviews for training and 25,000 for testing, with labels 0 or 1 indicating positive and negative reviews.
    \item \textbf{Reuters}~\cite{Reuters} is a dataset consisting of 11,228 newswires from Reuters, labeled over 46 topics. This was originally generated by parsing and preprocessing the classic Reuters-21578 dataset. Each newswire is encoded as a list of word indexes (integers). 
\end{itemize}
 
\subsection{Baseline Algorithms}
Since MASE can be treated as a constructed perturbation-based method for explanations, we choose the two most popular perturbation-based interpreting approaches (LIME~\cite{ribeiro2016should}, SHAP~\cite{lundberg2017unified}) as its competitors. In addition, we also introduce a random assignment of importance as a baseline to visualize the improvement of these methods.
These baseline algorithms are briefly summarized below. 
\begin{itemize}
\item \textbf{Random} assigns uniformly distributed random order for each word. 
Any effective explanation model should perform better than random rank since it doesn't unitize any information of the model. 
\item \textbf{LIME} is proposed by Ribeiro et al.~\cite{ribeiro2016should}, an algorithm that can explain the predictions of classifier or regressor by approximating it locally with an interpretable model. In LIME implementation, the number of features was set to 80, and the sample size was set to 1000.
\item \textbf{SHAP} is proposed by Scott Lundberg et al.~\cite{lundberg2017unified}, a game-theoretic approach that can explain any machine learning model using the classic Shapley values. We adopted the implementation of Deep SHAP, a faster (but only approximate) algorithm to compute SHAP values for deep learning models that are based on connections between SHAP and the DeepLIFT algorithm. We exclude this method from BERT since it is model-specific and is not supported in the original paper.
\item \textbf{PI}~\cite{altmann2010permutation} stands for permutation importance, which is based on repeated permutations of the input vector for estimating the measured importance for each word in a non-informative setting.
\item \textbf{Grad} is gradient-based explanation method. Since symbolic inputs (e.g., words) are often represented as one-hot vectors $x_t \in \{0, 1\}^{|V|}$ and embedded via a real-valued matrix $M$ such that $e_t = M\vv{x}_t$. Gradients are computed with respect to the individual
entry of embedding $E = [e_1, \dots, e_{|X|}]$. We use the $L^2$ norm, proposed by Poerner et al.~\cite{poerner2018evaluating} and Hechtlinger et al.~\cite{hechtlinger2016interpretation}, to reduce vectors of gradients to a single value.
\end{itemize}

\subsection{Evaluation Metrics}
Motivated by previous work~\cite{zhou2021evaluating, jacovi2020towards}, we measure the performance of the proposed method by faithfulness. ``Faithfulness'' refers to how accurately the explanation method reflects the true reasoning process of the target model~\cite{jacovi2020towards}. Specifically, we use delta accuracy to measure faithfulness/explainability. Delta accuracy is defined as follows:
\begin{equation}
    \begin{split}
\text { Delta Accuracy } &=\frac{|CC-MC|}{CC}
    \end{split}
    \label{eq:delta}
\end{equation}
where CC represents the correctly classified inputs in classification tasks without masking, and MC is the misclassified inputs after masking the important words specified by the explaining methods. 

This design is based on the fact that gradually masking the ``most positive'' words in positive content makes the text approximate to negative since negative words are retained; similarly, gradually masking the ``most negative'' words in a negative text makes the text tends to more positive since positive words are reserved. Based on Equation~\ref{eq:delta}, a reasonable explaining method is expected to be able to detect the most important features in the input, achieving a large delta accuracy value.



{\setlength\doublerulesep{0.5pt}
\begin{table*}[!h]
\small
\centering
\resizebox{\linewidth}{!}{
\begin{tabular}{c|cccc|cccc}
\toprule[1pt] \midrule
\multicolumn{1}{c|}{} &\multicolumn{4}{c|}{\textbf{LSTM-IMDB}} &\multicolumn{4}{c}{\textbf{LSTM-Reuters}} \\ \hline
&\textbf{Top-1} &\textbf{Top-5} &\textbf{Top-10} &\textbf{Top-15} &\textbf{Top-1}  &\textbf{Top-5}  &\textbf{Top-10}  &\textbf{Top-15} \\   
\midrule
\textbf{Random} &0.015$\pm$0.012 &0.063$\pm$0.012 &0.116$\pm$0.015 &0.158$\pm$0.021 &0.018$\pm$0.005 &0.051$\pm$0.016 &0.099$\pm$0.035 &0.163$\pm$0.054 \\ 
\textbf{LIME} &0.071$\pm$0.022 &0.113$\pm$0.032 &0.175$\pm$0.027 &0.214$\pm$0.020 &0.016$\pm$0.022 &0.073$\pm$0.026 &0.123$\pm$0.031 &0.192$\pm$0.043 \\ 
\textbf{SHAP}  &\textbf{0.178$\pm$0.014 }&\textbf{0.663$\pm$0.022} &\textbf{0.895$\pm$0.039 }&\textbf{0.969$\pm$0.041} &0.028$\pm$0.018 &0.078$\pm$0.025 &0.150$\pm$0.026 &0.229$\pm$0.033 \\
\textbf{PI} &0.053$\pm$0.018 &0.103$\pm$0.033 &0.153$\pm$0.034 &0.191$\pm$0.036 &0.018$\pm$0.013 &0.068$\pm$0.029 &0.119$\pm$0.028 &0.157$\pm$0.043 \\
\textbf{Grad*} &0.036$\pm$0.017 &0.163$\pm$0.026 &0.258$\pm$0.030 &0.367$\pm$0.034 &0.088$\pm$0.018 &0.198$\pm$0.024 &0.266$\pm$0.025 &0.328$\pm$0.017 \\ 
\textbf{MASE} &0.174$\pm$0.018 &0.646$\pm$0.054 &0.817$\pm$0.050 &0.876$\pm$0.055 &0.095$\pm$0.018 &0.481$\pm$0.030 &\textbf{0.732$\pm$0.034 }&\textbf{0.841$\pm$0.047} \\
\textbf{MASE\_Sparse} &0.171$\pm$0.020 &0.643$\pm$0.037 &0.818$\pm$0.051 &0.866$\pm$0.035 &\textbf{0.104$\pm$0.021 }&\textbf{0.483$\pm$0.033 }&0.646$\pm$0.038 &0.693$\pm$0.039 \\ 
\bottomrule[1pt]
\end{tabular}}
\caption{Delta accuracy of explaining LSTM on IMDB and Reuters Dataset by different masking schemes.}  
\label{tab:performance1}
\vspace{-2mm}
\end{table*}


\subsection{Setup}
In our experiment, we utilized a pre-trained LSTM classifier (trained on the IMDB and Reuters datasets) and a pre-trained BERT classifier (trained on IMDB) as the deep learner to be explained. We compared both MASE and MASE\_Sparse with three baselines: two popular model-agnostic methods - SHAP and LIME, and a random setting. All local explanations were computed by 1000 model evaluations. For the proposed MASE and MASE\_Sparse, the perturbation was based on the embedding results, while the LIME and SHAP models utilized a word-level perturbation scheme, where a word is either perturbed or not with a given probability.

{\setlength\doublerulesep{0.8pt}
\begin{table*}[!h]
\small
\centering
\resizebox{\linewidth}{!}{ 
\begin{tabular}{c|cccc|cccc}
\toprule[1pt] \midrule
\multicolumn{1}{c|}{} &\multicolumn{4}{c|}{\textbf{BERT-IMDB}}  &\multicolumn{4}{c}{\textbf{BERT-Reuters}} \\ \hline
&\textbf{Top-1} &\textbf{Top-5} &\textbf{Top-10} &\textbf{Top-15} &\textbf{Top-1}  &\textbf{Top-5}  &\textbf{Top-10}  &\textbf{Top-15} \\   
\midrule
\textbf{Random} &0.007$\pm$0.005 &0.023$\pm$0.017 &0.039$\pm$0.028 &0.047$\pm$0.025 &0.011$\pm$0.017 &0.020$\pm$0.015 &0.034$\pm$0.024 &0.049$\pm$0.022 \\ 
\textbf{LIME} &0.045$\pm$0.022 &0.055$\pm$0.032 &0.061$\pm$0.027 &0.062$\pm$0.020 &0.041$\pm$0.027 &0.053$\pm$0.029 &0.064$\pm$0.023 &0.063$\pm$0.021 \\ 
\textbf{SHAP} &0.033$\pm$0.010 &0.117$\pm$0.018 &0.171$\pm$0.016 &0.219$\pm$0.030 &0.030$\pm$0.014 &0.226$\pm$0.020 &0.284$\pm$0.019 &0.356$\pm$0.029 \\ 
\textbf{PI} &0.009$\pm$0.013 &0.032$\pm$0.017 &0.047$\pm$0.013 &0.050$\pm$0.029 &0.026$\pm$0.031 &0.049$\pm$0.034 &0.056$\pm$0.043 &0.060$\pm$0.051 \\ 
\textbf{Grad*}  &0.024$\pm$0.009 &0.097$\pm$0.014 &0.129$\pm$0.012 &0.147$\pm$0.010 &0.018$\pm$0.023 &0.067$\pm$0.026 &0.075$\pm$0.031 &0.083$\pm$0.029 \\
\textbf{MASE} &0.052$\pm$0.018 &0.216$\pm$0.054 &\textbf{0.322$\pm$0.050} &\textbf{0.396$\pm$0.055 }&\textbf{0.061$\pm$0.025 }&\textbf{0.256$\pm$0.026 }&0.358$\pm$0.032 &\textbf{0.409$\pm$0.033} \\ 
\textbf{MASE\_Sparse}  &\textbf{0.054$\pm$0.020 }&\textbf{0.230$\pm$0.037} &\textbf{0.322$\pm$0.051 }&0.390$\pm$0.035 &0.058$\pm$0.024 &0.251$\pm$0.029 &\textbf{0.362$\pm$0.039} &0.402$\pm$0.036 \\
\bottomrule[1pt]
\end{tabular}}
\caption{Delta accuracy of explaining BERT-based classifier on IMDB and Reuters Dataset by different masking schemes. }  
\label{tab:performance2}
\vspace{-2mm}
\end{table*}

\begin{table*}[t]
\small
\centering
\resizebox{\linewidth}{!}{ 
\begin{tabular}{c|cccc|cccc}
\hline
\multicolumn{1}{c|}{} &\multicolumn{4}{c|}{\textbf{BERT-IMDB}}  &\multicolumn{4}{c}{\textbf{BERT-Reuters}} \\ \hline
&\textbf{Top-1} &\textbf{Top-5} &\textbf{Top-10} &\textbf{Top-15} &\textbf{Top-1}  &\textbf{Top-5}  &\textbf{Top-10}  &\textbf{Top-15} \\   
\hline
\textbf{Random} &0.007$\pm$0.005 &0.023$\pm$0.017 &0.039$\pm$0.028 &0.047$\pm$0.025 &0.011$\pm$0.017 &0.020$\pm$0.015 &0.034$\pm$0.024 &0.049$\pm$0.022 \\ 
\textbf{LIME} &0.045$\pm$0.022 &0.055$\pm$0.032 &0.061$\pm$0.027 &0.062$\pm$0.020 &0.041$\pm$0.027 &0.053$\pm$0.029 &0.064$\pm$0.023 &0.063$\pm$0.021 \\ 
\textbf{SHAP} &0.033$\pm$0.010 &0.117$\pm$0.018 &0.171$\pm$0.016 &0.219$\pm$0.030 &0.030$\pm$0.014 &0.226$\pm$0.020 &0.284$\pm$0.019 &0.356$\pm$0.029 \\ 
\textbf{PI} &0.009$\pm$0.013 &0.032$\pm$0.017 &0.047$\pm$0.013 &0.050$\pm$0.029 &0.026$\pm$0.031 &0.049$\pm$0.034 &0.056$\pm$0.043 &0.060$\pm$0.051 \\ 
\textbf{Grad*} &0.024$\pm$0.009 &0.097$\pm$0.014 &0.129$\pm$0.012 &0.147$\pm$0.010 &0.018$\pm$0.023 &0.067$\pm$0.026 &0.075$\pm$0.031 &0.083$\pm$0.029 \\
\textbf{MASE} &0.052$\pm$0.018 &0.216$\pm$0.054 &\textbf{0.322$\pm$0.050} &\textbf{0.396$\pm$0.055} &\textbf{0.061$\pm$0.025} &\textbf{0.256$\pm$0.026} &0.358$\pm$0.032 &\textbf{0.409$\pm$0.033} \\ 
\textbf{MASE\_Sparse} &\textbf{0.054$\pm$0.020} &\textbf{0.230$\pm$0.037} &\textbf{0.322$\pm$0.051} &0.390$\pm$0.035 &0.058$\pm$0.024 &0.251$\pm$0.029 &\textbf{0.362$\pm$0.039} &0.402$\pm$0.036 \\
\hline
\end{tabular}
}
\caption{Delta accuracy of explaining BERT-based classifier on IMDB and Reuters datasets by different masking schemes.}  
\label{tab:performance2}
\end{table*}

\paragraph{\textbf{Training Details}} We built the LSTM from scratch and used the publicly available pre-trained BERT model~\cite{turc2019} weights to initialize the BERT classifier. The LSTMs were fine-tuned to have a testing accuracy of $81.0\%$ on the IMDB dataset and $84.2\%$ on the Reuters dataset, while the BERT-based classifier was fine-tuned to achieve a testing accuracy of $89.3\%$ on the IMDB dataset. For each baseline, we customized the original method to fit utilizing the fine-tuned classifiers.

\begin{figure*}[!t]
    \includegraphics[width=\linewidth]{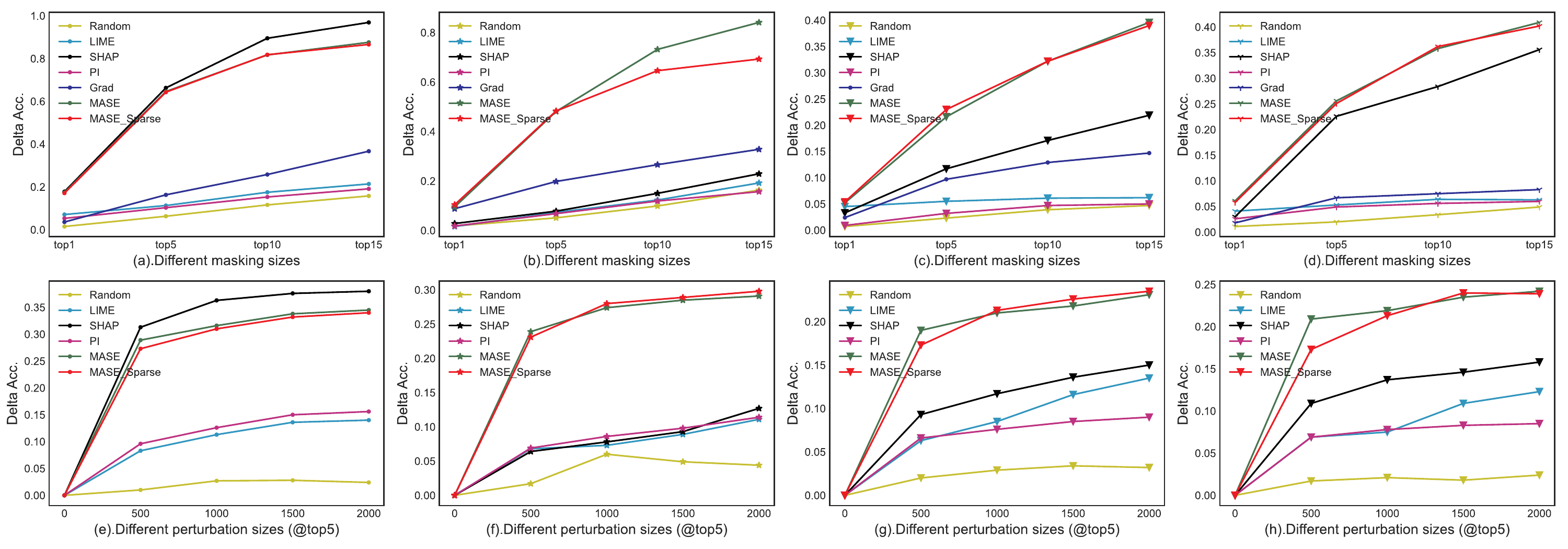}
    \caption{Experimental results of investigating the effectiveness of embedding-based estimation, different masking sizes, and perturbation sizes. Column 1, column 2, and column 3 are conducted on the IMDB dataset with the LSTM model, the Reuters dataset with the LSTM model, and IMDB dataset with the BERT-based model, respectively.}  
  \label{fig:result_nlp}
\vspace{-3mm}
\end{figure*}

The test dataset was randomly selected from the original dataset, with a sample size of 1000 for each batch, and had already been correctly classified by the target model.
After masking the top-k words that have the highest scores of the explanations, we computed the accuracy of the target model on the masked data. Ideally, consistent and trustworthy interpretation methods should be able to identify important words within the sentences, thus producing the largest accuracy drop with the top-k masking scheme. The results are summarized in Table~\ref{tab:performance1} and Table~\ref{tab:performance2}. 

Table~\ref{tab:performance1} shows the performance of explaining LSTM models on the IMDB and Reuters datasets, measured in terms of delta accuracy in different top-k value masking schemes. The experiment is based on explaining correctly classified inputs, thus the classification accuracy is 1 before masking. As we can see, PI slightly outperforms the Random method. MASE and MASE\_Sparse deliver a comparatively good performance close to that of SHAP on the IMDB dataset, while LIME has a poor performance measured by delta accuracy in comparison with SHAP, Grad, and MASE on the dataset. Interestingly, when tested on the Reuters dataset, MASE and MASE\_Sparse outperformed both LIME, Grad, PI, Grad, and SHAP significantly. Overall, SHAP achieved the best delta accuracy for the IMDB dataset but underperformed on the Reuters dataset, while MASE delivered more consistent results on both test datasets. 
Specifically, when masking the word with the largest feature importance in the Reuters dataset, MASE achieved the best performance with a delta accuracy of 0.095, which indicates that 9.5\% of the testing data is incorrectly classified due to information loss. 
When the top 10 words were masked, MASE achieved the best performance on the Reuters dataset, with a delta accuracy of 73.2\%, while MASE\_Sparse had the best performance on the Reuters dataset when top-5 words are masked, with a delta accuracy of 48.3\%. When more important words were masked, the proportion of data classified as the opposite of its ground truth increased, substantially increasing the delta accuracy. This rationale is consistent with the experimental results. When the top 15 words were masked, MASE achieved the best performance on the Reuters datasets. To sum up, MASE demonstrated its competitiveness compared to the leading model-agnostic method (SHAP), PI, and model-specific method Grad.

Table~\ref{tab:performance2} shows the performance of explaining BERT-based classifiers on the IMDB dataset and Reuters dataset, measured in terms of the delta accuracy achieved for different top-k value masking schemes. As we can see, MASE and MASE\_Sparse consistently outperformed SHAP, LIME, Random, PI, and Grad, with the best delta accuracy of 5.4\% when masking the top 1 word, an accuracy of 23.0\% when masking the top 5 words, an accuracy of 32.2\% when masking the top 10 words, and an accuracy of 39.6\% when masking the top 15 words. Comparing these results with those presented in Table~\ref{tab:performance1}, masking the most important features in the explaining BERT resulted in a slower decrease in delta accuracy while masking the most important features in the explaining LSTMs led to a faster decrease in delta accuracy. This implies that the predictions from the BERT-based classifier rely on a larger number of keywords than those needed by the LSTMs.
In addition, this observation is consistent with the fact that BERT is inherently more robust to masking since it randomly masks features during the training process~\cite{devlin2018bert}.
The experimental results from Table~\ref{tab:performance1} and Table~\ref{tab:performance2} clearly demonstrate that MASE and MASE\_Sparse achieve the most stable performance when the underlying model becomes complex.

\subsubsection{Ablation Study}
We further conduct ablation studies to investigate the effectiveness of the proposed MASE framework. The experimental results are shown in Figure~\ref{fig:result_nlp}. From sub-figures (a), (b), (c), and (d) in Figure~\ref{fig:result_nlp}, we can tell that increasing the masking size and masking important features (words) will result in a large delta accuracy, which is sensible since removing/masking important features drives the underlying deep model to have opposite classification result. From sub-figures (e), (f), (g), and (h), we see that increasing perturbation size will decrease delta accuracy. This effect from perturbation size is obvious in the first 1,500 perturbations. However, as we further increase the perturbation size, the marginal contribution decreases. Moreover, comparing LIME (PI/Random) with MASE (embedding-based perturbation), we conclude that the embedding-based estimator is more effective since it results in a better performance (large delta accuracy).



\section{Conclusion}
In this paper, we present MASE, a statistically consistent and model-agnostic saliency estimation framework for explaining deep learning models in NLP settings. 
MASE utilizes a constrained Normalized Linear Gaussian Perturbation (NLGP) on the embedding representations to generate meaningful perturbations for the text models instead of the human heuristic. We illustrate that the variant of MASE can further provide a sparse and efficient solution. The proposed MASE framework was thoroughly evaluated by explaining LSTM and BERT-based models on real-world datasets.
The experimental results demonstrate that our proposed MASE framework 
evidently outperforms the currently widely used model-agnostic methods, LIME, PI, and SHAP. In addition, the explanation check confirms that our method is faithful to the target models and provides an alternative way to visualize the learning procedure of the deep models utilizing saliency explanations. 

\section{Acknowledgement}
This work was supported by the U.S. National Science Foundation under Grant \textbf{NSF IIS-2238700.} The views and conclusions contained in this paper are those of the authors and should not be interpreted as representing any funding agencies.

\bibliographystyle{IEEEtran}
\bibliography{reference}

\end{document}